\documentclass[letterpaper, 10 pt, conference]{ieeeconf}  % Comment this line out if you need a4paper

\IEEEoverridecommandlockouts                              % This command is only needed if 
\usepackage{graphicx} % for pdf, bitmapped graphics files
\usepackage{algorithm}
\usepackage{mathtools}
\usepackage[noend]{algpseudocode}
\usepackage{subcaption}
\usepackage{amsfonts}
\usepackage[noadjust]{cite}
\usepackage{caption}
\makeatletter
\newcommand\fs@betterruled{%
  \def\@fs@cfont{\bfseries}\let\@fs@capt\floatc@ruled
  \def\@fs@pre{\vspace*{5pt}\hrule height.8pt depth0pt \kern2pt}%
  \def\@fs@post{\kern2pt\hrule\relax}%
  \def\@fs@mid{\kern2pt\hrule\kern2pt}%
  \let\@fs@iftopcapt\iftrue}
\floatstyle{betterruled}
\restylefloat{algorithm}
\makeatother

\title{\LARGE \bf
Dot-to-Dot: Explainable Hierarchical Reinforcement Learning \\for Robotic Manipulation
}

\author{Benjamin Beyret$^{1}$, Ali Shafti$^{1}$ and A. Aldo Faisal$^{1,2,3,4}$% <-this % stops a space
\thanks{$^{1}$Brain and Behaviour Lab: Dept. of Computing, $^{2}$ Dept. of Bioengineering, $^{3}$ Data Science Institute,  $^{4}$ UKRI Centre for Doctoral Training in AI for Healthcare -- {\tt bb1010@ic.ac.uk}, {\tt a.shafti@ic.ac.uk}, {\tt a.faisal@ic.ac.uk}}}%

\begin{document}

\maketitle
\thispagestyle{empty}
\pagestyle{empty}

%%%%%%%%%%%%%%%%%%%%%%%%%%%%%%%%%%%%%%%%%%%%%%%%%%%%%%%%%%%%%%%%%%%%%%%%%%%%%%%%
\begin{abstract}

Robotic systems are ever more capable of automation and fulfilment of complex tasks, particularly with reliance on recent advances in intelligent systems, deep learning and artificial intelligence. However, as robots and humans come closer in their interactions, the matter of interpretability, or explainability of robot decision-making processes for the human grows in importance. A successful interaction and collaboration will only take place through mutual understanding of underlying representations of the environment and the task at hand. This is currently a challenge in deep learning systems. We present a hierarchical deep reinforcement learning system, consisting of a low-level agent handling the large actions/states space of a robotic system efficiently, by following the directives of a high-level agent which is learning the high-level dynamics of the environment and task. This high-level agent forms a representation of the world and task at hand that is interpretable for a human operator. The method, which we call Dot-to-Dot, is tested on a MuJoCo-based model of the Fetch Robotics Manipulator, as well as a Shadow Hand, to test its performance. Results show efficient learning of complex actions/states spaces by the low-level agent, and an interpretable representation of the task and decision-making process learned by the high-level agent.
% A video of the trained agents can be found here: \url{https://youtu.be/o94PjPqf5HA}

\end{abstract}

%%%%%%%%%%%%%%%%%%%%%%%%%%%%%%%%%%%%%%%%%%%%%%%%%%%%%%%%%%%%%%%%%%%%%%%%%%%%%%%%
\section{Introduction}

Robots are increasingly present in our lives, from production lines to homes, hospitals and schools, we rely more and more on their presence. Robots working directly with humans are either controlled directly by human input \cite{Tanaka}, pre-programmed to follow a pre-planned choreography of movements between the human and the robot \cite{bauer2008human} or programmed to follow a control law. Having an intelligent robot that can learn a task and adapt to situations unseen before, while interacting with a human, is a major challenge in the field. 

The rapid advance of artificial intelligence has led researchers to the creation of intuitive agents representing robots in simulated environments, or on real-world robots in specific use cases. The majority of these intelligent systems are based on deep learning \cite{Kober-2013,levine2016end}. While these systems produce impressive results in automation, they also result in what is referred to as a ``blackbox" algorithm, i.e. the decision-making process, and factors affecting it are not clear to the human user. A human interacting with such a system will not realise how their actions are being interpreted by the robotic system, and how they lead to robot actions. These systems are therefore not explainable. For a human interacting with such a system, not knowing how the robot's actions relate to their own behaviour or the environment, can lead to reduced trust, reliability, safety and efficiency for the overall interaction. 

Reinforcement learning is one of the promising solutions to the intelligent robotics problem \cite{guenter2007reinforcement,gu_2017}. Most of these techniques use some form of optimization to solve a task in a given robotics environment \cite{Kober-2013}, however very few actually look at the inherent structure of the tasks, or are concerned with creating higher level representations which are interpretable by an interacting human user.

\begin{figure}[t] 
    \centering
    \begin{subfigure}[b]{0.22\textwidth}
        \includegraphics[width=\textwidth]{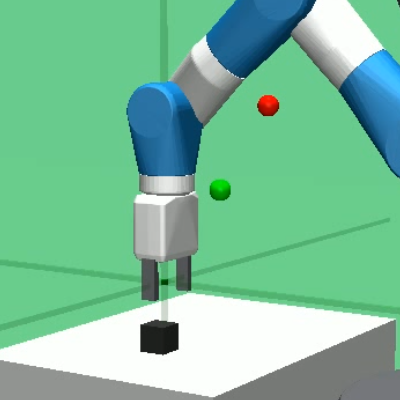}
        \caption{starting position}
        \label{fig:FetchPickAndPlaceMulti1}
    \end{subfigure}
    ~
    \begin{subfigure}[b]{0.22\textwidth}
        \includegraphics[width=\textwidth]{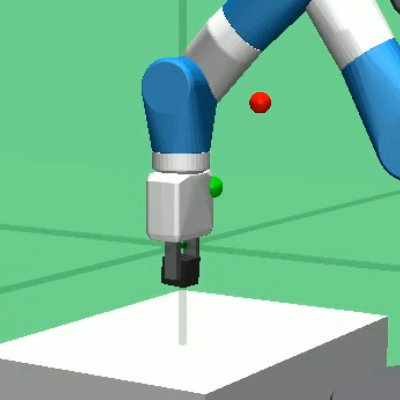}
        \caption{intermediate state}
        \label{fig:FetchPickAndPlaceMulti2}
    \end{subfigure}
    \\
    \begin{subfigure}[b]{0.22\textwidth}
        \includegraphics[width=\textwidth]{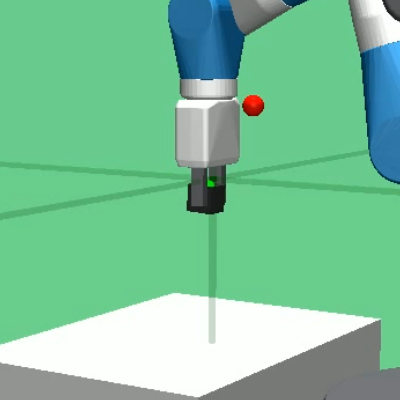}
        \caption{sub-goal reached}
        \label{fig:FetchPickAndPlaceMulti3}
    \end{subfigure}
        ~
    \begin{subfigure}[b]{0.22\textwidth}
        \includegraphics[width=\textwidth]{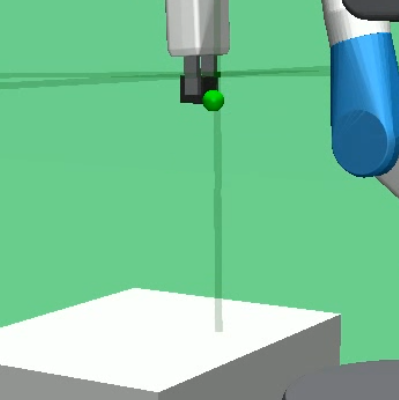}
        \caption{end goal reached}
        \label{fig:FetchPickAndPlaceMulti4}
    \end{subfigure}
    \caption{Dot-to-Dot in action: the agent learns to break down a complex task into simpler low-level actions. On \texttt{FetchPickAndPlace}: we first generates the green dot as a sub-goal, using the high-level policy (\ref{fig:FetchPickAndPlaceMulti1}), then the low-level policy reaches that sub-goal (\ref{fig:FetchPickAndPlaceMulti2} and \ref{fig:FetchPickAndPlaceMulti3}), and finally the end goal (\ref{fig:FetchPickAndPlaceMulti4})}\label{fig:FetchPickAndPlaceMulti}
\end{figure}

In this work, our main goal is to create an intelligent robotic agent that can solve manipulation tasks by learning some form of curriculum \cite{Bengio2009} as well as a structure of movement, in a manner that conserves interpretability for a human operator. Our solution has a hierarchical structure where a complex task is split into multiple low-level consecutive simpler actions. This idea builds on the concept of action grammars governing human behaviour, which we previously used in a human-robot interaction scenario \cite{shafti2019gaze}. In our proposed hierarchy, the high-level agent will divide the full task into smaller (easier) actions that a low-level agent can learn to fulfil. In this manner the low-level agent will learn to make sense of the many degrees of freedom of the system with a curriculum learning approach, while the high-level agent will learn the overall dynamics of the environment and the task at hand, creating a high-level representation governing the decision-making process. Effectively, the high-level agent serves as an interpreter between the human, the environment and the low-level agent controlling the robot's many degrees of freedom. This will serve as a fundamental step in making intelligent robots operating through reinforcement learning that is explainable to human users.

The paper is structured as follows: Section II covers the necessary background upon which our method is built. Section III presents our proposed algorithm, \emph{Dot-to-Dot}, with its design and implementation explained in detail. Section~IV reports the results of training and task performance for Dot-to-Dot, followed by a discussion on the inner representation the high-level agent creates of its world, and its interpretability. Finally, Section V concludes the paper and covers potential future work.

\section{Background}
In this section, we detail the building blocks of our work. Our algorithm relies on three existing algorithms and concepts, namely Deep Deterministic Policy Gradients (DDPG, \cite{Lillicrap_al_2015}), Hindsight Experience Replay (HER, \cite{Andrychowicz_2017}) and Hierarchical Reinforcement Learning \cite{Barto2003}, which we describe below. 

Let us first define a few notations used throughout this paper: an observation (i.e. states) space $\mathcal{O}$, an action space $\mathcal{A}$ and a set of goals $\mathcal{G} \subseteq \mathcal{O}$. For example, in the case of a robotic arm that needs to push a cube around a table, $\mathcal{O}$ could be the position of the gripper and position of the cube, $\mathcal{G}$ can be a position on the table where the cube needs to be moved to, and $\mathcal{A}$ can be a set of actions that end up moving the gripper. We define $g\in\mathcal{G}$ as the goal of an episode, sampled at $t=0$ when resetting an environment, $ag \in \mathcal{G}$ an achieved goal at time $t>0$ (e.g. position of the cube at time $t$), and $sg \in \mathcal{G}$ a defined sub-goal which will be a waypoint to $g$. Finally, $r_t$ is the reward at time $t$ and $R_t=\sum_{k=0}^t r_k$ the cumulative reward obtained by the agent at time $t$.

\subsection{Deep Deterministic Policy Gradients}
The Deep Deterministic Policy Gradients (DDPG) method \cite{Lillicrap_al_2015} combines several techniques together in order to solve continuous control tasks. DDPG is an extension of the Deterministic Policy Gradients (DPG) algorithm \cite{Silver_al_2014}. DPG uses an actor-critic architecture \cite{Sutton_1998} where the actor's role is to take actions in the environment while the critic is in charge of assessing how useful these actions are to complete a given task. In other terms, both the actor's and critic's policies are updated using the temporal difference error of the critic's value function.  Moreover, DPG uses a parameterized function $\mu(s|\theta^{\mu})$ as the actor and $Q(s,a)$ as the critic function, which is learned through Q-learning. The updates of the actor consist in applying to both functions the gradients of the expected return $J = \mathbb{E} [R_t]$ with respect to the parameters $\theta^{\mu}$.

Finally, Deep DPG extends DPG by using neural networks as function approximators for $\mu(s|\theta^{\mu})$ and $Q(s,a)$, implementing an architecture similar to  \cite{Mnih_al_2015}.

\subsection{Hindsight Experience Replay}
In Hindsight Experience Replay (HER), Andrychowicz et al. \cite{Andrychowicz_2017} use an elegant trick to train agents faster in large states and actions spaces. Observing that very few traces actually yield a positive outcome (i.e. goal reached), the authors propose to make use of every trace no matter whether the goal was reached or not. In fact, they note that regardless of what the objective of a series of actions is, and no matter the outcome, we can still acquire valuable information from experience; i.e. we can still learn how every state of a trace was reached by looking at the previous states visited and actions taken in those states. During an episode, HER samples a batch of $N$ traces $(s^i_t, ag^i_t, a^i_t, s^i_{t+1}, ag^i_{t+1}, r^i_t, g^i)$ for $t<T$  and $i \leq N$. Then, at training,  for a proportion $K$ of these traces, HER will replace $g$ with a randomly selected $ag^i_{t^{\prime}}$ with $t^{\prime} \in \mathcal{U}([t+1; T])$, $\mathcal{U}$ being the uniform distribution; meaning that it assumes the incorrect state we ended up in, was in fact our goal. Therefore, in hindsight, we look at goals that were achieved instead of the original goal, learning from mistakes. This technique proved to be greatly successful for diverse robotics tasks in simulation \cite{Andrychowicz_2017}.

\subsection{Hierarchical Reinforcement Learning}

Hierarchical reinforcement learning \cite{Barto2003} is a technique that intends to address the problem of planning how to solve a task, by introducing various levels of abstraction. Most of the time it involves several policies, which interact with each other, often as a high-level policy dictating which of another set of policies to use and when. One such structure \cite{Barto2003} involves several low-level policies called options, which each learn how to complete a specific task, while a high-level policy decides which low-level option to use when. Another approach \cite{dayan1993feudal,Vezhnevets_2017} is to match observations to goals using a higher-level policy commanding a lower-level policy, each with its own level of abstraction and temporal resolution.

Our contribution builds on top of these techniques, combining them to create an agent that achieves structured robotic manipulation. Closely related work is that of Nachum et al. \cite{nachum2018} where a similar hierarchical structure is used to solve exploration tasks in a data-efficient manner and using observations as goals. In \cite{Levy2018} another hierarchical structure is used to solve common toy environment tasks. The latter uses different low-level policies for each sub-goals while the former focuses on exploration tasks. In our work on the other hand, we focus on the inherent structure of robotic manipulation and reusing low-level skills across different steps of a task, as well as explainability.

\section{Design and Implementation}

\subsection{Design}
\begin{figure}[tp]
    \centering
    \begin{subfigure}[b]{0.45\textwidth}
        \includegraphics[width=\textwidth]{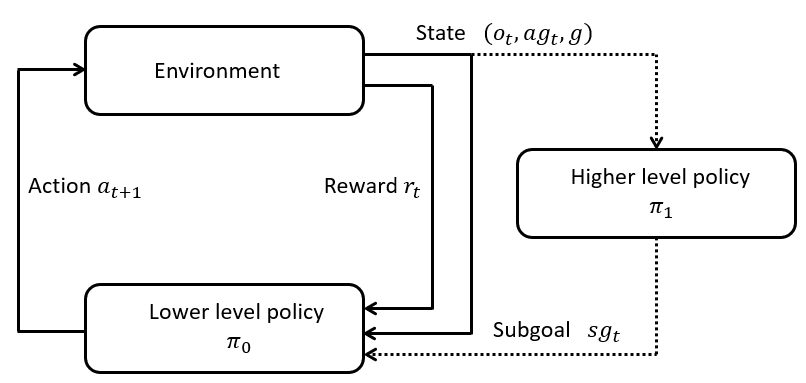}
        \caption{Dot-to-Dot structure. Sub-goals generated by $\pi_1$, fed to $\pi_0$}
        \label{flowchart}
    \end{subfigure}

    \begin{subfigure}[b]{0.48\textwidth}
        \includegraphics[width=\textwidth]{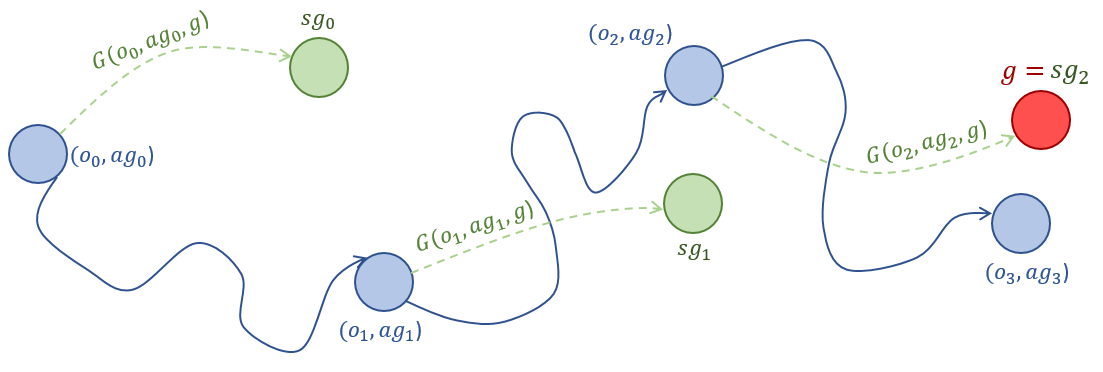}
        \caption{Example of an episode to train DtD}
        \label{walk}
    \end{subfigure}
% \vspace{-4mm}
\caption{Dot-to-Dot setup and example of an episode during training}
\end{figure}

We introduce a hierarchical reinforcement learning architecture which we call Dot-to-Dot (DtD). This comes from the fact that our architecture is made of a high-level agent that generates sub-goals for a low-level agent, which follows them one by one to achieve the high-level task -- resembling the children's game of the same name, where connecting dots creates an overall drawing. \newline
To implement this, we define two policies: a low-level one $\pi_0: \mathcal{O} \times \mathcal{G} \rightarrow \mathcal{A}$ and a high-level one $\pi_1: \mathcal{O} \times \mathcal{G} \rightarrow \mathcal{G}$. The low-level policy $\pi_0$ is trained using DDPG, that takes as inputs observations as well as goals generated by the high-level policy $\pi_1$. This is described in figure \ref{flowchart}.

To teach our agent complex sequences of action in a sparse reward setup, we note that it is easier to learn how to reach nearby goals than further ones. This defines the DtD method: in order to train the low-level agent in an efficient manner, for a given starting point $(o_0, ag_0, g) \in \mathcal{O}\times\mathcal{G}\times\mathcal{G}$ in an episode, $\pi_1$ (i.e. high-level agent) first generates sub-goals that are in the vicinity of $ag_0$. This is done by making the first sub-goal $sg_1$ be a noisy perturbation of $ag_0$, formally: $sg_0 = ag_0+\mathcal{N}(0,\sigma)$, with $\mathcal{N}$ a Gaussian distribution and $\sigma$ the noise parameter, effectively ignoring the final goal $g$. Doing so, the low-level policy  $\pi_0$ can be trained easily on reaching these newly defined sub-goals. The other central idea of DtD is to ignore whether or not goals and sub-goals have been achieved during an episode, and to use HER for both $\pi_0$ and $\pi_1$ to extract as much information as possible from past experience.

Let us take an example of a trace and describe the training process. Figure \ref{walk} shows an episode with two sub-goals.
In this example observations are represented in blue ($o_i$), sub-goals ($sg_i$) in green and the final goal ($g$) in red; $ag_i$ refers to achieved goals. First, we use $\pi_1$ to generate $sg_0$. As mentioned above, at first $sg_0$ is a noisy perturbation of $ag_0$ such that: $sg_0 = ag_0+\mathcal{N}(0,\sigma)$. Then $\pi_0$ tries to reach $sg_0$ in a certain number of steps and reaches $ag_1$. At the beginning of training, we mostly have $ag_1 \neq sg_0$, which is fine, as we ignore this and generate a new sub-goal $sg_1$ such that $sg_1 = ag_1+\mathcal{N}(0,\sigma)$. Again, $\pi_0$ generates actions to reach $sg_1$ but instead reaches $ag_2$. Finally, as we want our algorithm to learn which traces are useful for reaching a given goal, and which are not, we constrain the last sub-goal to be equal to the actual final goal $g$. Doing so, we can learn if a sequence of actions and sub-goals can reach a given goal or not. \newline
Once this episode is done, we have obtained a series of states ($o_i$) reached by $\pi_0$, sub-goals ($sg_i$) generated by $\pi_1$ and actually achieved goals ($ag_i$). In order to train our agents, we use DDPG in combination with HER. The low level agent is therefore easier to train, as it needs to reach goals closer to its current state. 
% as it only has to learn to push the cube from where it is to a nearby location. 
The interesting part is the training of the high level agent, resulting in $\pi_1$: as HER allows for replacement of the actual goal $g$ with some achieved goal $ag$, it makes training more efficient by learning which sub-goal can make $\pi_0$ reach which goal. \newline
This method of training intrinsically generates a form of curriculum learning as the overall agent will learn more and more complex tasks along the way. In fact, it will first explore its surroundings, then learn how to generate useful sub-goals for a given goal and finally put both together to solve tasks.

\subsection{Implementation}

For our experiments we used three robotic environments from OpenAI gym robotics suite \cite{gym}: \texttt{FetchPush}, \texttt{FetchPickAndPlace} and \texttt{HandManipulateBlock}. These are simulations based on MuJoCo \cite{Todorov_2012}. All of them are goal oriented setups, in both \texttt{Fetch} environments the agent is a robotic arm based on the Fetch robot \cite{fetch} which must move a black cube to a desired goal represented by a red dot which is either on a table (\texttt{FetchPush}) or in the air (\texttt{FetchPickAndPlace}). For both, the actions are 4-dimensional with 3 continuously controlling the position of the gripper and one for opening/closing the gripper. The observations are the Cartesian positions of the gripper and the object as well as both their linear and angular velocities. In the \texttt{FetchPush} environment however, the gripper is always set to be closed, which forces the agent to push the cube around, making the task rely heavily on physical properties of the block and table (i.e. friction), which need to be learned by the agent. The \texttt{Hand} environment is a Shadow hand \cite{kochan_2005} holding a cube which must be manipulated in-hand to reach a  desired orientation goal; note that in figure \ref{fig:HandBlockMulti}, the goal and sub-goal orientations are displayed to the right of the hand. In the  \texttt{HandManipulateBlock}, the agent directly controls the individual joints of the hand, which makes for a much more challenging task. In this case, the actions are 20-dimensional (24 DoF, 4 of which are coupled), and the observations are the position and velocities of the 24 joints, as well as the object's Cartesian position, its rotation as a quaternion as well as linear and angular velocities \cite{plappert_2018}. 

Algorithm \ref{alg:sub-goals} presents more details on DtD. Note that the sub-goal selection agent $\pi_1$ is in charge of initially generating sub-goals as local noisy perturbations of the current achieved goal, and gradually handing over to the actor-critic agent for sub-goals inference. This can also be done in an $\epsilon$-greedy manner.

\begin{algorithm}[t]
    \caption{Dot-to-Dot (DtD)}
    \label{alg:sub-goals}
    \begin{algorithmic}
    \State Initialize DDPG agent $\pi_0$ (low-level) and HER buffer $\mathcal{R}$
    \State Initialize sub-goal selection agent $\pi_1$
    \For{epoch\,=\,$0$,\,$N_{epochs}$}
    	\For{episode\,=\,$1$,\,$M$}
        	\State Reset the environment to obtain $(s_0, ag_0, g)$
          	\State $s=s_0$, $ag=ag_0$ 
           	\For{$n\,=\,0,\,N$} \Comment{N=number of sub-episodes}
            	\State Sample $sg \leftarrow \pi_1(s,g)$ 
               	\For{$t\,=\,T_n,\,T_{n+1}$}
                	\State Sample an action $a_t \leftarrow \pi_0(s || sg)$ 
                	
                	\Comment{$||$ denotes concatenation}
                    \State Execute $a_t$, observe $(s_{t+1}, ag_{t+1}, r_t)$
                    \State $s=s_t$, $ag=ag_t$
               	\EndFor
                \For{$t\,=\,0,\,T_N$}
                	\State $(s_t,ag_t,sg_t,a_t,r_t,s_{t+1},ag_{t+1},g) \xrightarrow{\text{store}} R$
              	\EndFor
        	\EndFor
    	\EndFor
      	\For{training\,=\,$0$,\,$N_{trainings}$}
        	\State perform an update of $\pi_0$ using $\mathcal{R}$
            \State perform an update of $\pi_1$ using $\mathcal{R}$
        \EndFor
    \EndFor
    \end{algorithmic}
\end{algorithm}

There are a few technical details that we need to address before looking at the actual implementation and the results we obtained. First of all, we made the choice to always force the last sub-goal of an episode to match the actual end-goal set at the beginning of the episode. Another possibility is to fully ignore the actual goal every time we explore, and only set sub-goals as defined above. However, experiments have shown this technique to be quite inefficient due to the fact that the agent does not even try reaching the goal and we believe this leads to inefficient training of the policy as a result. Another choice we made is that of replacing goals with achieved goals during training. This results in applying HER to sub-goals when training $\pi_1$'s network. 
In practice, consider an example with 5 sub-goals. During training we sample an episode stored in our replay buffer $\mathcal{R}$, from this episode we extract: $(sg_0,sg_1,sg_2,sg_3,sg_4)$ the sub-goals, $(o_0,o_1,o_2,o_3,o_4)$ and $(ag_0,ag_1,ag_2,ag_3,ag_4)$ the corresponding observations and achieved goals, and $g$ the goal. In classic experience replay training we would train the network on traces such as $(sg_2,o_2,g)$. Instead, similarly to HER, we choose to replace $g$ with a later achieved goal $ag$ for a certain proportion of traces. For example, we could replace $g$ with $ag_3$ and train on $(sg_2,o_2,ag_3)$ instead, virtually making the trace successful as $ag_3$ was reached by definition. This proved very efficient while training and will be used in all following experiments.

\section{Results and Analysis}

\subsection{Training performance}

\begin{figure}[t]
    \centering
    \begin{subfigure}[b]{0.44\textwidth}
        \includegraphics[width=\textwidth]{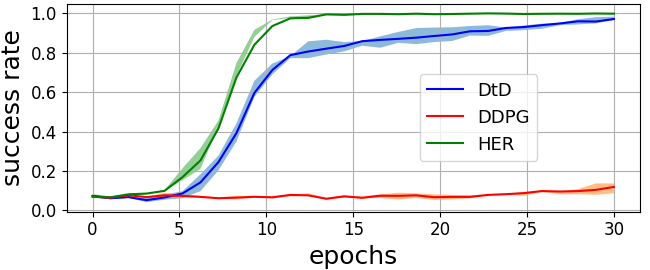}
        \caption{Success rates during training for \texttt{FetchPush}}
        \label{fig:trainCurvePush}
    \end{subfigure}

    \begin{subfigure}[b]{0.44\textwidth}
        \includegraphics[width=\textwidth]{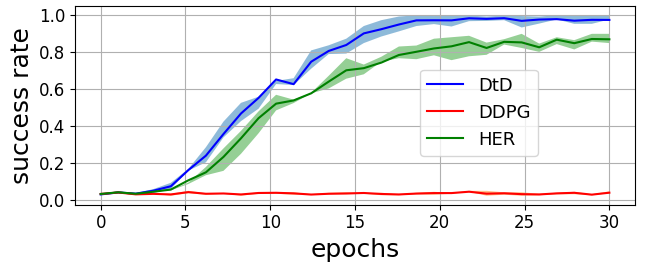}
        \caption{Success rates during training for \texttt{FetchPickAndPlace}}
        \label{fig:trainCurvePick}
    \end{subfigure}

    \caption{Training curves for DDPG, HER and DtD agents on the \texttt{Fetch} tasks. The curves represent the mean and the confidence intervals are $25^{th}$ to $75^{th}$ percentiles.}   \label{fig:trainCurves}
\end{figure}

Figures \ref{fig:trainCurvePush} and \ref{fig:trainCurvePick} show the evolution of the success rate of DDPG, HER and DtD over epochs, on the \texttt{FetchPush} and \texttt{FetchPickAndPlace} tasks respectively. We note that DtD is marginally slower than HER at training on the easier task \texttt{FetchPush} while being marginally faster on the more difficult one \texttt{FetchPickAndPlace}. Meanwhile, DDPG did not succeed at the tasks in the given number of epochs. Training was done on five random seeds, the figures show the mean and confidence intervals on these seeds.

\vspace{-2mm}

\subsection{Task performance}
\begin{figure}[bp]
    \centering
    \begin{subfigure}[b]{0.14\textwidth}
        \includegraphics[width=\textwidth]{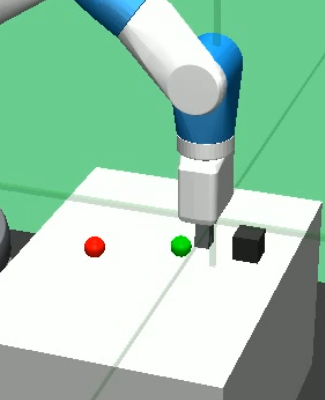}
        \caption{starting position}
        \label{fig:FetchPushMulti1}
    \end{subfigure}
    ~
    \begin{subfigure}[b]{0.14\textwidth}
        \includegraphics[width=\textwidth]{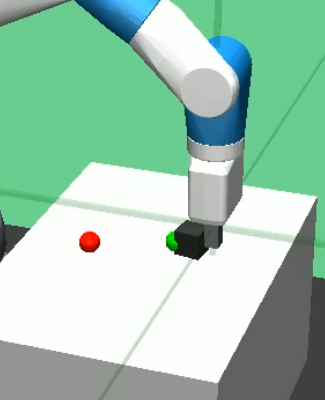}
        \caption{reach sub-goal}
        \label{fig:FetchPushMulti2}
    \end{subfigure}
    ~
    \begin{subfigure}[b]{0.14\textwidth}
        \includegraphics[width=\textwidth]{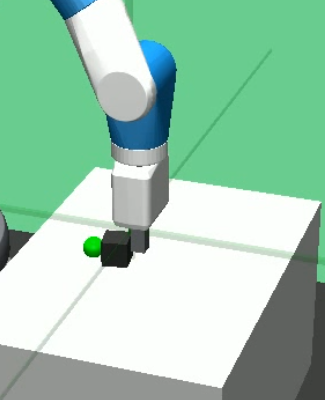}
        \caption{goal reached}
        \label{fig:FetchPushMulti3}
    \end{subfigure}
    \caption{DtD agent on \texttt{FetchPush}: the agent first generates the green dot as a sub-goal using the high-level policy (\ref{fig:FetchPushMulti1}), then the low-level agent reaches that sub-goal (\ref{fig:FetchPushMulti2}), and finally the last sub-goal which is now equal to the end goal (\ref{fig:FetchPushMulti3}) }   \label{fig:FetchPushMulti}
\end{figure}

In this section we present a few still frames of episodes using the best policy obtained after training. We are interested in assessing whether or not the sub-goals generated by the high-level agent are meaningful and make sense in terms of positioning. \newline
We first present the results obtained on the \texttt{Fetch} environments. where the end goal is represented by a red dot, while sub-goals are represented by a green dot. We look at configurations with only one sub-goal as the table is rather small, note that the last sub-goal is forced to be the end-goal which is why the red dot turns green in the last frames (Figures \ref{fig:FetchPushMulti3} and \ref{fig:FetchPickAndPlaceMulti4}).

In both environments, figures \ref{fig:FetchPushMulti1} and \ref{fig:FetchPickAndPlaceMulti1} show the sub-goals generated are indeed located approximately in the middle of the initial cube's position and end-goal. We can also see that the low-level agent does indeed succeed in reaching all sub-goals and completes the task. The \texttt{FetchPush} example also shows that the agent managed to learn some form of concept and representation of its environment, the most obvious one being the fact that sub-goals have to be generated on the tabletop. In fact, during training, the high-level agent is not constrained at all in terms of sub-goal generation. As mentioned in the previous part, sub-goals can be generated anywhere in the vicinity of the initial cube's location, and therefore can even appear inside the table or in the air. However, traces that include these types of sub-goals will bear very low rewards and therefore force the agent to generate sub-goals close to the tabletop. 

\begin{figure}[t!]
        \centering
        \begin{subfigure}[b]{0.2\textwidth}
            \includegraphics[width=\textwidth]{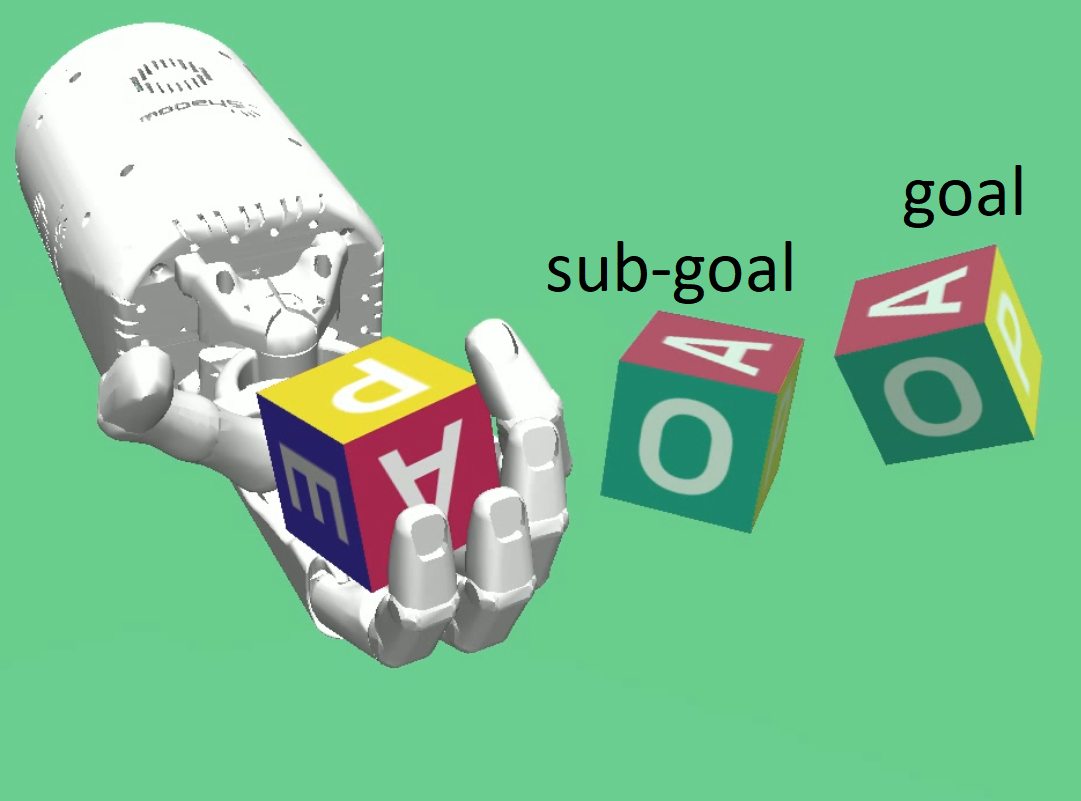}
            \caption{starting position}
            \label{fig:HandBlockMulti1}
        \end{subfigure}
        ~
        \begin{subfigure}[b]{0.2\textwidth}
            \includegraphics[width=\textwidth]{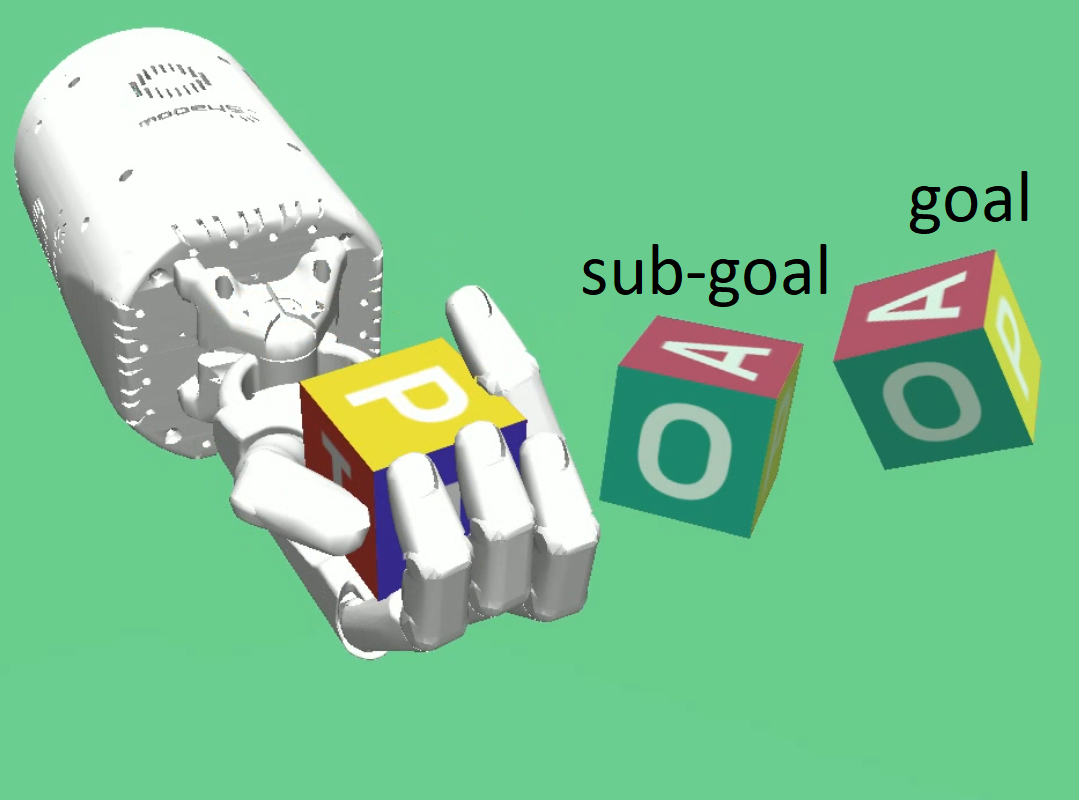}
            \caption{targets sub-goal}
            \label{fig:HandBlockMulti2}
        \end{subfigure}
        \\
        \begin{subfigure}[b]{0.2\textwidth}
            \includegraphics[width=\textwidth]{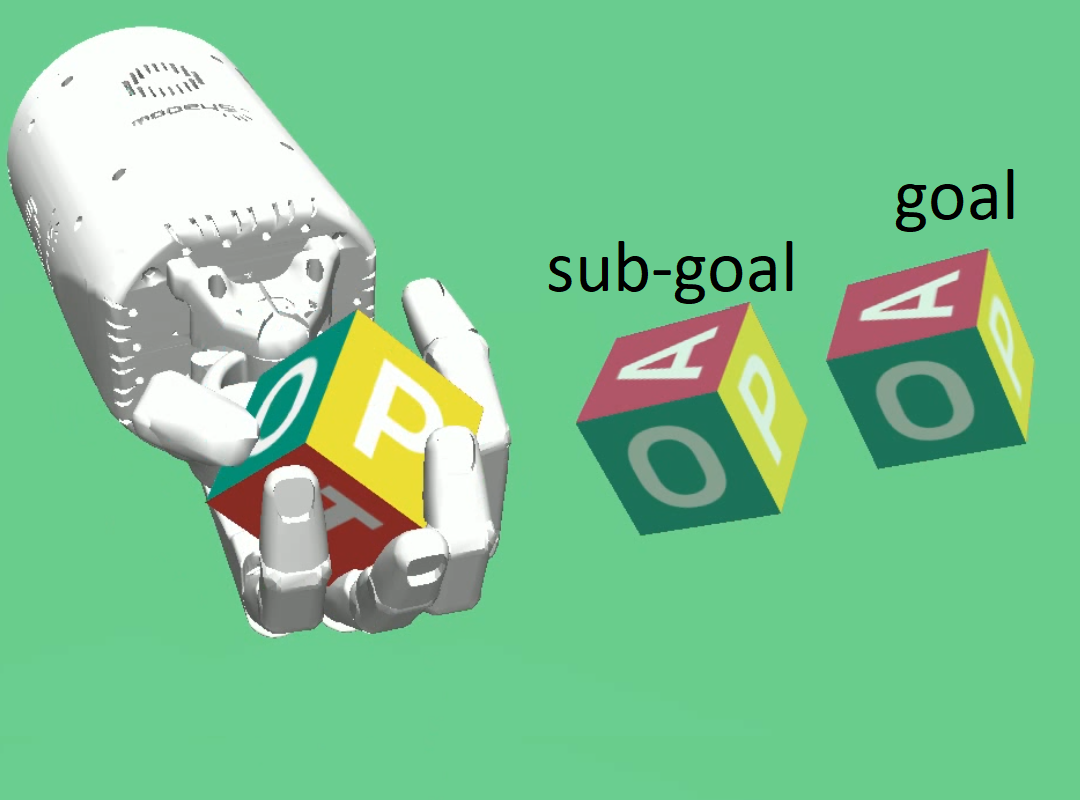}
            \caption{sub-goal missed}
            \label{fig:HandBlockMulti3}
        \end{subfigure}
            ~
        \begin{subfigure}[b]{0.2\textwidth}
            \includegraphics[width=\textwidth]{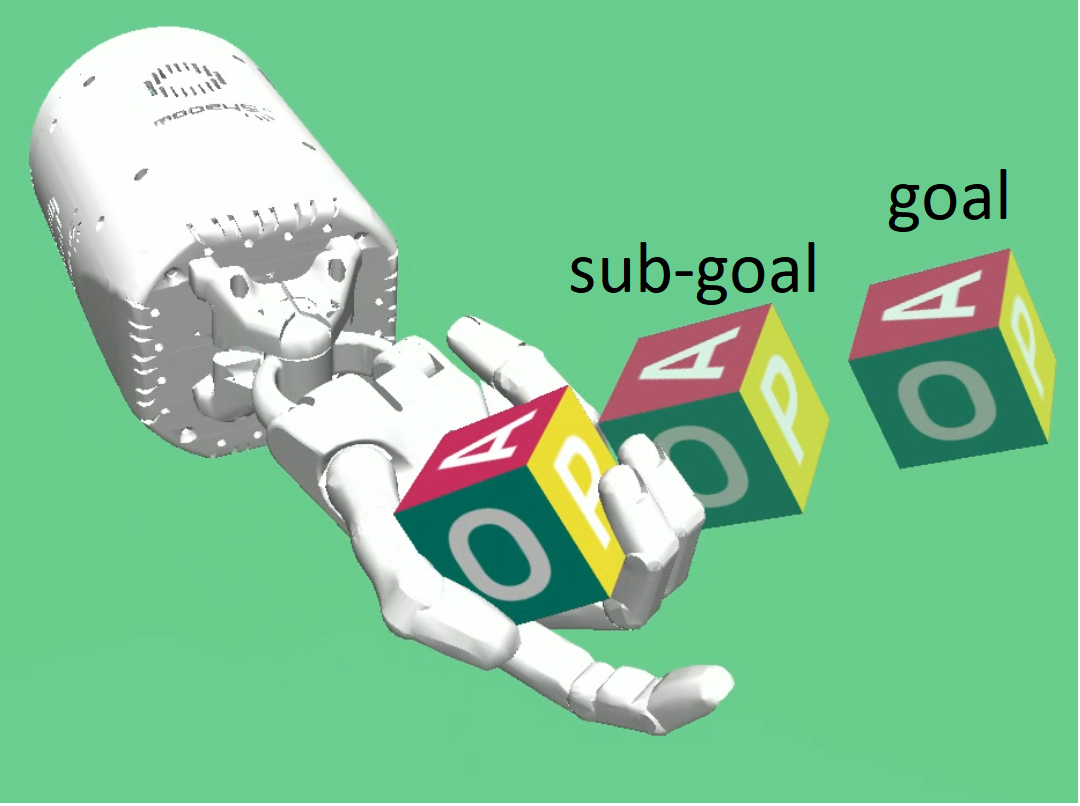}
            \caption{end goal reached}
            \label{fig:HandBlockMulti4}
        \end{subfigure}
        \caption{ DtD agent on the \texttt{HandManipulateBlock} environment. The agent starts with the cube in an initial position (\ref{fig:HandBlockMulti1}), rotates the cube to reach the first sub-goal (\ref{fig:HandBlockMulti2}) but doesn't manage to reach the sub-goal in the first sub-episode as the sub-goal shifts to the end goal (\ref{fig:HandBlockMulti3}), this does not prevent the agent from reaching the end goal as the sub-goal generated was  useful to the low level agent (\ref{fig:HandBlockMulti4})}\label{fig:HandBlockMulti}
    \end{figure}

Figure \ref{fig:HandBlockMulti} shows frames of an episode on the ShadowHand simulation where a cube must be rotated to the goal orientation. We can again see that the sub-goals are generated to be on the way to the end-goal, however it is harder to observe than in the \texttt{Fetch} environments due to the nature of the task. As figure \ref{fig:HandBlockMulti3} shows, the agent did not manage to reach the first sub-goal in the given time. Despite this miss, we can see that the agent positioned the cube closer to the target anyway, as shown by the yellow side being positioned correctly. Therefore, even though the low-level agent may sometimes miss a sub-goal due to time constraints, the generated sub-goals help reach an end-goal, as shown in Figure \ref{fig:HandBlockMulti4}.

\subsection{High-level agent's inner representation}
In this part, we look at the way the high-level agent values different regions of the environment as candidates for the low-level agent's sub-goals. This allows us to interpret the way the agent represents the various environments internally, and makes it easier for humans to read into the decision-making process of the agent, improving explainability. We use a specific configuration of the \texttt{FetchPush} environment, where the initial position and the goal have been chosen to be at opposite corners of the table. The idea is to look at the table from above with the robot north of the table, and discretize both its $X$ and $Y$ axes. We then define sub-goals as pairs of the discrete axes: $sg = (x,y)$. Finally, for each of these sub-goals, we compute $\pi_1$'s $Q$-values $Q_{\pi_1}(o,sg,g)$ which is the expected value of choosing  $sg$ as a sub-goal in the given configuration. A low value means that the sub-goal is not a good candidate, and the overall episode will yield a low cumulated reward.

\begin{figure}[b!]
    \centering
    \begin{subfigure}[b]{0.4\textwidth}
        \includegraphics[width=\textwidth]{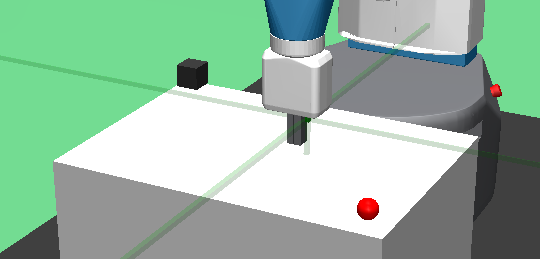}
        \caption{Setup 1}
        \vskip 0.1in
        \label{fig:heatmapSetup1}
    \end{subfigure}

    \begin{subfigure}[b]{0.4\textwidth}
        \includegraphics[width=\textwidth]{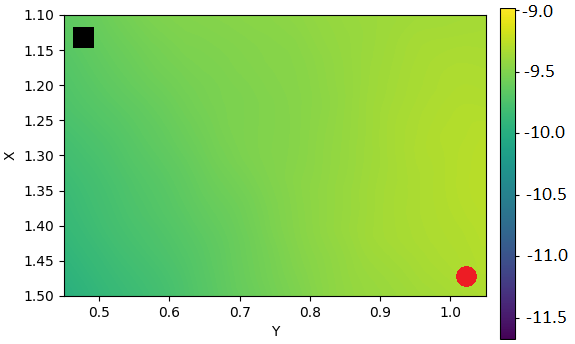}
        \vskip -0.1in
        \caption{$\pi_1$'s Q-function after one epoch}
        \vskip 0.1in
        \label{fig:heatmapBefore1}
    \end{subfigure}
 
    \begin{subfigure}[b]{0.4\textwidth}
        \includegraphics[width=\textwidth]{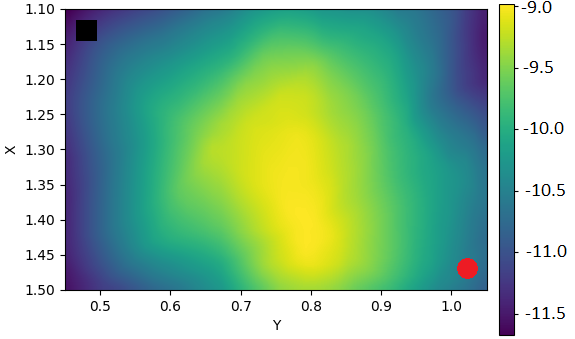}
        \vskip -0.1in
        \caption{$\pi_1$'s Q-function after training}
        \label{fig:heatmapAfter1}
    \end{subfigure}
    \caption{Setup with initial state and goal diagonally opposed on the table. The heatmaps show the highest areas (yellow) of value for the high-level agent to predict a sub-goal. The black square represents the position of the cube, the red circle is the end goal}\label{fig:heatmapsDiagonal1}
\end{figure}
In figure \ref{fig:heatmapsDiagonal1}, we show the two extremes in terms of distance from the initial state to the end-goal. The \texttt{FetchPush} environment is ideal to look at inner representations as it is almost two dimensional (goals are always on the tabletop). Figure \ref{fig:heatmapSetup1} shows this setup. We first look at the $Q$-values at the very beginning of training, after just one epoch, as shown in figure \ref{fig:heatmapBefore1}. We can see the values are very close to each other and spread in a small interval, which shows that the high-level agent does not have a clear representation of the environment yet. Despite this lack of clear representation, the agent seems to attribute higher values to sub-goals closer to the end-goal (to the right of the table) rather than those close to the starting position (to its left). This makes sense as sub-goals close to the end-goal are most likely to allow the agent to reach its destination, and are therefore the very first sub-goals that lead to successful traces. \newline
Finally, after training, the best policy's $Q$-values are represented in figure \ref{fig:heatmapAfter1}. We can now see that the values are spread over a much larger interval, and therefore the higher values mean the associated sub-goals are clearly better candidates. This area of higher value is located well in the middle between the starting position and the end-goal. Note that in practice, the sub-goal will only be the point with the highest value on this heatmap. We can therefore conclude that the agent learnt a good representation of its environment as well as a notion of distance, considering the effects of friction and the dynamics involved in pushing a block around to an end-goal.

\section{CONCLUSIONS}

We set out to create an agent that can learn to complete tasks in environments that are challenging by nature: high dimensional and only presenting sparse rewards. We also aimed at finding a structure that equips the agent with the ability to create a representation of its environment that can be easily understood by humans. We achieved this by combining several techniques to produce the Dot-to-Dot algorithm, learning a hierarchical structure of motion and manipulation through curriculum learning. This was implemented and tested through OpenAI Gym and MuJoCo, with the Fetch Robotics Manipulator and the Shadow Hand environments.

In terms of training times we obtained results equivalent to the current baselines, however we have shown that on top of this, we achieved to provide the agent with the ability to produce interpretable representations of its environment. The agent learnt a notion of distance, being able to create waypoints to an end-goal, splitting a complex task into several easier consecutive ones and reusing learnt behaviour across these. We believe this can serve as a fundamental first step to help make robotic agents intelligent while preserving the explainability of their actions.

Future work will focus on improving exploration for sub-goals in the vicinity of a current position, one solution for this could be to use intrinsic motivation and curiosity \cite{oudeyer2016,Colas2018b}. Another lead could be to produce more goals that do not necessarily need to be achieved, leading the agent towards a direction instead of having waypoints. Finally, we are interested in testing the algorithm on a real robotic system. This method complements the work we presented in \cite{shafti2019gaze}, making that robotic setup a good candidate for real-world use of Dot-to-Dot.
% \foaddtolength{\textheight}{-here12cm}   % This command serves to balance the column lengths

\bibliographystyle{bib/IEEEtran}
\bibliography{bib/IEEEexample}

\end{document}